\pdfoutput=1

\documentclass[11pt]{article}

\usepackage[preprint]{acl}

\usepackage{times}
\usepackage{latexsym}
\usepackage{multirow}
\usepackage{hyperref}
\usepackage{array}
\usepackage{wrapfig}
\usepackage{makecell}
\usepackage{booktabs}
\usepackage{xspace}
\usepackage{xcolor}
\usepackage{amssymb} 
\usepackage[affil-it]{authblk}

\usepackage[T1]{fontenc}

\usepackage[utf8]{inputenc}

\usepackage{microtype}

\usepackage{inconsolata}

\usepackage{graphicx}

\newcommand{\ourmethod}{\textsc{Abacus-SQL}\xspace}

\title{\ourmethod: A Text-to-SQL System Empowering Cross-Domain and Open-Domain Database Retrieval}

\author{
Keyan Xu, Dingzirui Wang, Xuanliang Zhang, Qingfu Zhu, Wanxiang Che\thanks{Corresponding author.} \\
Harbin Institute of Technology \\
\{kyxu, dzrwang, xuanliangzhang, qfzhu, car\}@ir.hit.edu.cn
}

\begin{document}
\maketitle
\begin{abstract}
The existing text-to-SQL systems have made significant progress in SQL query generation, but they still face numerous challenges. Existing systems often lack retrieval capabilities for open-domain databases, requiring users to manually filter relevant databases. Additionally, their cross-domain transferability is limited, making it challenging to accommodate diverse query requirements. To address these issues, we propose \ourmethod. \ourmethod utilizes database retrieval technology to accurately locate the required databases in an open-domain database environment. It also enhances the system cross-domain transfer ability through data augmentation methods. Moreover, \ourmethod employs Pre-SQL and Self-debug methods, thereby enhancing the accuracy of SQL queries. Experimental results demonstrate that \ourmethod performs excellently in multi-turn text-to-SQL tasks, effectively validating the approach's effectiveness. \ourmethod is publicly accessible at 
\href{https://huozi.8wss.com/abacus-sql/}{https://huozi.8wss.com/abacus-sql/}.
\end{abstract}

\begin{table*}[t]
    \centering
    \begin{tabular}{lccc}
\toprule
\textbf{System} & \textbf{Multi-Turn?} & \textbf{Database Retrieval?} & \textbf{Cross-Domain?}\\
\midrule
DBGPT \cite{xue_db-gpt_2024} & \textcolor{red}{\(\times\)} & \textcolor{red}{\(\times\)} & \textcolor{red}{\(\times\)}\\
PHOTON \cite{zeng_photon_2020} & \textcolor{red}{\(\times\)} & \textcolor{red}{\(\times\)} & \textcolor{red}{\(\times\)}\\
SQLChat\footnotemark[1] & \textcolor{red}{\(\times\)} & \textcolor{red}{\(\times\)} & \textcolor{red}{\(\times\)}\\
Vanna\footnotemark[2] & \textcolor{green}{\(\checkmark\)} & \textcolor{red}{\(\times\)} & \textcolor{red}{\(\times\)}\\
WrenAI\footnotemark[3] & \textcolor{green}{\(\checkmark\)} & \textcolor{red}{\(\times\)} & \textcolor{red}{\(\times\)}\\
\midrule
\ourmethod & \textcolor{green}{\(\checkmark\)} & \textcolor{green}{\(\checkmark\)} & \textcolor{green}{\(\checkmark\)}\\
\bottomrule
    \end{tabular}
    \caption{Comparison of \ourmethod with previous systems.}
    \label{tab:system}
\end{table*}

\section{Introduction}
Text-to-SQL \cite{yu_spider_2019} is a natural language processing (NLP) technique designed to automatically convert natural language queries into SQL statements, thereby lowering the barrier to data querying. This technique has been widely applied in areas such as business analytics and customer support \cite{liu_survey_2024, hong_next-generation_2024, katsogiannis-meimarakis_survey_2023}. However, existing text-to-SQL technologies remain challenging to use due to complex database structures, ambiguous natural language understanding, and diverse user query habits \cite{xue_db-gpt_2024}. To improve usability, it is essential to develop a powerful, intuitive and user-friendly text-to-SQL system capable of accurately interpreting users’ diverse natural language queries and generating efficient and precise SQL statements.

Previous text-to-SQL systems \cite{zeng_photon_2020, zeng_n-best_2023} have demonstrated the potential of natural language interaction with databases, with notable innovations from systems such as DB-GPT \cite{xue_db-gpt_2024} and PHOTON \cite{zeng_photon_2020}. DB-GPT possesses powerful SQL generation capabilities, while its novel Retrieval-Augmented Generation (RAG) knowledge system and adaptive learning mechanism further enhance query efficiency. PHOTON enhances the system ability to handle ambiguous and complex user inputs by integrating deep learning with a human-in-the-loop correction mechanism, thereby improving its cross-domain adaptability and robustness.

Although existing text-to-SQL systems have made significant progress in SQL query generation, they still face several limitations (Table \ref{tab:system}). Current systems lack efficient \textbf{database retrieval capability} and struggle to automatically locate the required database in open-domain database environments, forcing users to manually filter databases, which reduces the system’s generality and efficiency. Additionally, existing systems exhibit limited \textbf{cross-domain transferability}, as most require pretraining for specific domains. This constraint restricts their applicability across different domains, making it increasingly difficult to meet the query needs of specialized databases.

To address the above limitations of existing text-to-SQL systems, we develop \ourmethod, focusing on enhancing multi-database retrieval performance and cross-domain transferability while introducing several innovative methods to optimize SQL generation.
First, \ourmethod supports retrieval in open-domain databases by leveraging beam search and query rewriting to accurately locate the required database.
Second, \ourmethod exhibits robust cross-domain transferability by utilizing data augmentation methods to synthesize demonstrations based on domain-specific databases, enabling the system to quickly adapt to diverse domain requirements.
Moreover, \ourmethod integrates pre-SQL and self-debug methods, ensuring the generation of high-quality SQL even in complex query scenarios, thereby further enhancing the system’s practicality and reliability.

Overall, we develop \ourmethod, a robust text-to-SQL system designed for cross-domain and open-domain database environments. Our main contributions are as follows:
\begin{itemize}
    \item \textbf{Database retrieval capability:} To address the retrieval challenges in multi-database environments, \ourmethod employs open-domain database retrieval method, enabling efficient retrieval of relevant databases.
    \item \textbf{Cross-Domain Transferability:} To enhance cross-domain transferability, \ourmethod utilizes data augmentation methods to synthesize examples from domain-specific databases, significantly improving cross-domain adaptability.
    \item \textbf{System Optimization:} To improve the quality of SQL query generation, \ourmethod incorporates multiple innovative methods, significantly enhancing the accuracy of results.
\end{itemize}

\begin{figure*}[t]
  \centering
    \includegraphics[width=\textwidth]{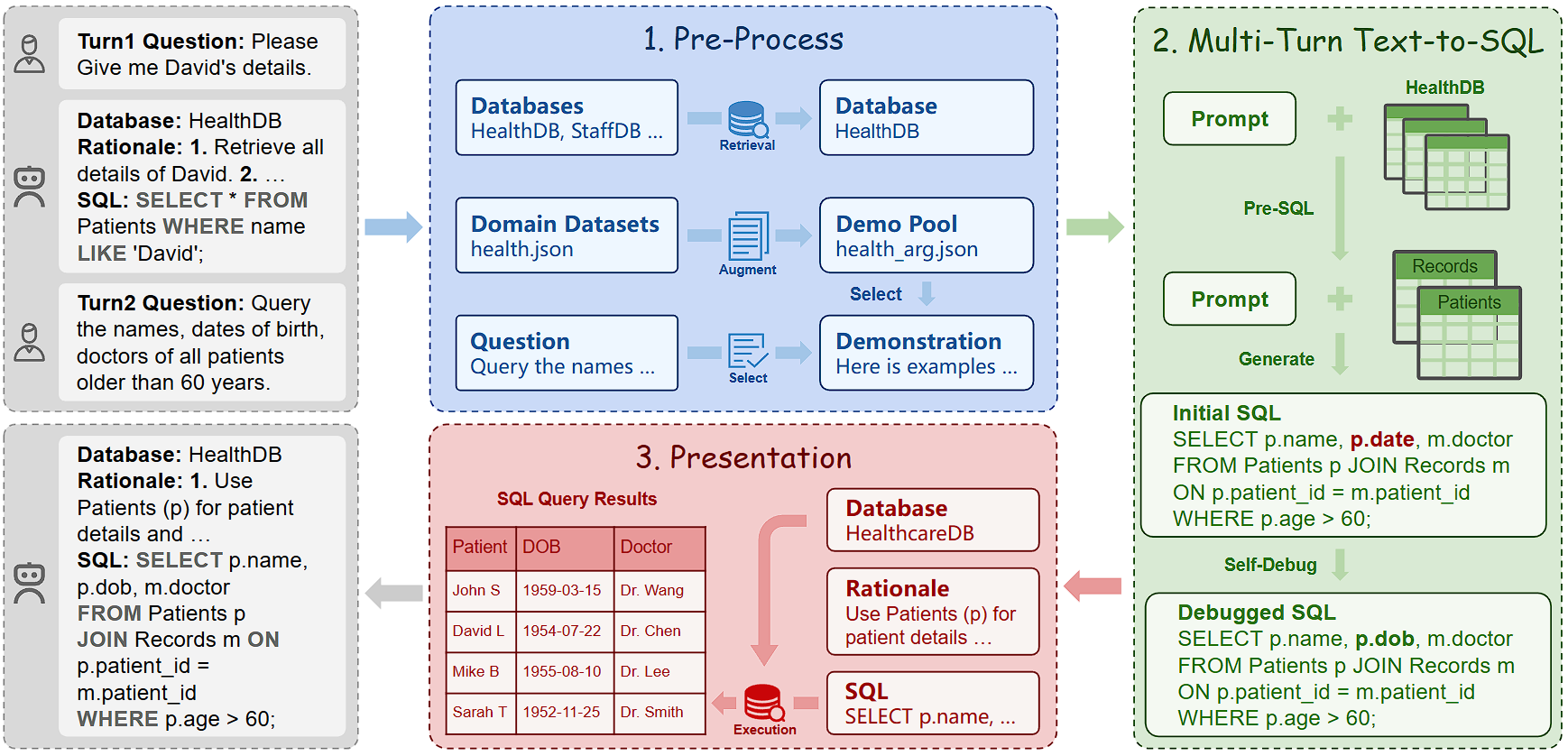}
    \caption{The illustration of \ourmethod, which consists of three steps: 
    \textit{1. Preprocessing}: Retrieves open-domain databases and enhances cross-domain transferability with data augmentation.
    \textit{2. Multi-turn Text-to-SQL}: Improves the accuracy of multi-turn SQL queries using Pre-SQL and Self-debug methods.
    \textit{3. Presentation}: Shows the inference process, SQL queries, and real-time execution results to users.}
    \label{fig:example}
\end{figure*}

\footnotetext[1]{\url{https://github.com/sqlchat/sqlchat}}
\footnotetext[2]{\url{https://github.com/vanna-ai/vanna}}
\footnotetext[3]{\url{https://github.com/Canner/WrenAI}}

\section{Related Work}
\subsection{Multi-turn Text-to-SQL}
Early multi-turn text-to-SQL research primarily relied on deep neural network models, improving SQL generation accuracy through specialized architectures. For example, \citet{wang_tracking_2020} proposed leveraging previous SQL queries to enhance parsing accuracy and contextual understanding, while RASAT \cite{qi_rasat_2022} introduced a relation-aware self-attention mechanism within the Transformer structure to improve dialogue context integration. However, such models face significant challenges including high data annotation costs and complex context management \cite{gao_text--sql_2023}.

With the advancement of large language models (LLMs), LLM-based methods have gradually become the mainstream, achieving high performance without additional fine-tuning, thereby reducing dependence on large datasets and computational resources \cite{hong_next-generation_2024}. ACT-SQL \cite{zhang_act-sql_2023} utilizes Chain-of-Thought reasoning to decompose multi-turn conversations into single-turn queries, handling dependencies through query rewriting and context completion. CoE-SQL \cite{zhang_coe-sql_2024} further optimizes this process by adopting an edit-based strategy that incrementally updates SQL queries, avoiding error accumulation caused by query rewriting, thereby improving stability and accuracy. Overall, the integration of LLMs has made multi-turn text-to-SQL more efficient and versatile, reducing resource demands while enhancing the coherence and precision of SQL generation \cite{zhang_coe-sql_2024}.

\subsection{Text-to-SQL System}
In recent years, text-to-SQL technology has made significant advancements, leading to the emergence of various open-source tools that simplify user-database interactions and enable non-expert users to easily access the data they need.
DB-GPT \cite{xue_db-gpt_2024} is a framework that integrates LLMs with database interaction technologies. It supports natural language queries, efficient SQL generation, multilingual support, and incorporates privacy protection and multi-agent collaboration strategies, offering new perspectives for text-to-SQL system development.
PHOTON \cite{zeng_photon_2020} is a cross-domain natural language interface database system that effectively enhances the handling of complex and ambiguous queries through deep learning and a human-in-the-loop correction mechanism.
SQLChat\footnotemark[1] adopts a conversational interaction model, enabling users to execute database operations through natural language.
WrenAI\footnotemark[2] functions as an SQL AI agent, supporting multi-database environments and integrating semantic understanding to improve query efficiency.
These tools have significantly driven the development of text-to-SQL, catering to diverse user needs and expanding the accessibility of database querying. 

However, existing systems often lack retrieval functionality for open-domain databases, increasing user operation complexity and time cost. They also struggle with cross-domain transferability, making it hard to adapt to different data structures and query needs. Therefore, enhancing the system's domain transferability and adaptability in multi-database environments is a key challenge for text-to-SQL systems.

\begin{figure*}[t]
  \centering
    \includegraphics[width=\textwidth]{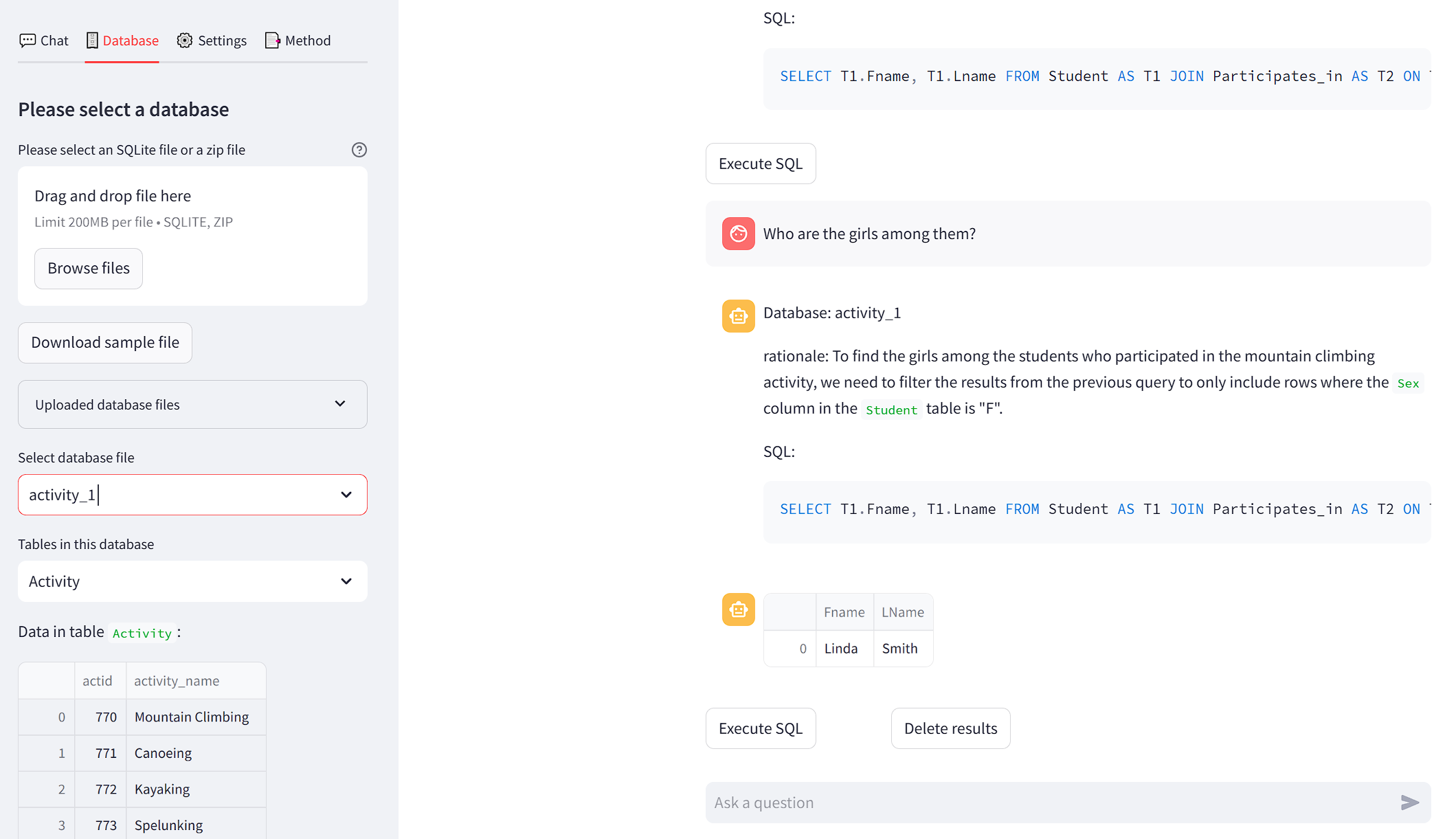}
    \caption{The interface of \ourmethod: The sidebar provides various functions, such as uploading and viewing user databases, as well as switching between sessions. The main area facilitates interaction with \ourmethod, allowing users to generate SQL queries and execute query results.}
    \label{fig:demo}
\end{figure*}

\section{System Workflow}
In this section, we introduce the workflow of our system, which is designed to address the limitations of previous systems, including insufficient retrieval capabilities, limited transferability, and suboptimal SQL generation. The workflow, as illustrated in the Figure \ref{fig:example}, consists of three core phases: preprocessing, multi-turn text-to-SQL, and presentation.
To overcome the shortcomings of existing systems, we implemented several optimizations. First, we employ the Murre method (Section \ref{sec:murre}) for automatic retrieval to extract databases relevant to the given query. Second, we utilize the fused method (Section \ref{sec:fused}) for data augmentation, enhancing the system’s cross-domain transferability. Finally, in the SQL generation phase, we introduce Pre-SQL (Section \ref{sec:presql}) and Self-debug (Section \ref{sec:debug}) to improve the accuracy of SQL generation.

\subsection{Preprocess}
During the initial data preprocessing stage, we prepare for subsequent SQL generation through three key steps: open-Domain database retrieval, augmentation and selection of demonstration, and extraction of database schema information.
\subsubsection{Open-Domain Database Retrieval}
\label{sec:murre}
We first automatically identify and select the most relevant database based on the user's query and the uploaded databases. This process consists of two steps: database matching, which aligns the user query with database schemas and metadata to determine databases likely containing the target information; and database prioritization, which evaluates and ranks multiple relevant databases to select the most suitable one. Specifically, we employ the Murre method from \cite{zhang_multi-hop_2024}, iteratively performing database beam scarch and query-related field elimination. Detailed implementation can be found in Appendix \ref{sec:appendix_murre}.

\subsubsection{Demonstration Selection}
\label{sec:fused}
We select demonstrations \cite{dong_survey_2024} from domain-specific datasets to help the model better align with domain characteristics for effective domain adaptation while also providing reference examples. We first employ data augmentation to expand either the default domain dataset or user-provided domain dataset, enhancing data diversity and adaptability to improve the model’s cross-domain transferability. Here, we adopt the Fused method from \cite{wang_improving_2024}, leveraging a large language model (LLM) to iteratively update the demonstration pool (detailed implementation is provided in Appendix \ref{sec:appendix_fused}). Subsequently, we perform demonstration selection using the BM25 \cite{robertson_probabilistic_2009} algorithm, which incorporates user query requirements, database schema, and dialogue context to retrieve demonstration from the predefined demonstration pool, providing valuable references for SQL generation.

\subsubsection{Schema Extraction}
Here, we systematically extract table schema from the previously selected database and precisely align the database structure with the user query. First, we retrieve table names, column names, data types, and their underlying relationships from the database and organize them into a format that is easily interpretable by the LLMs. Then, by aligning the fields in the user query with the database content, we ensure that the model accurately identifies the query intent, enabling the generated SQL to correctly map to the relevant tables and fields.

\begin{table*}[t]
\centering
\begin{tabular}{ll|cccc}
\toprule
\multirow{2}{*}{\textbf{Dataset}} & \multirow{2}{*}{\textbf{Method}} & \multicolumn{2}{c}{\textbf{7B}} & \multicolumn{2}{c}{\textbf{32B}} \\
 &  & \textbf{QEX} & \textbf{IEX} & \textbf{QEX} & \textbf{IEX} \\
 \midrule
\multirow{2}{*}{\textbf{Chase-C}} & Qwen2.5-Coder & $40.4$ & $11.1$ & $46.5$ & $18.0$ \\
 & \quad + \ourmethod & $\textbf{45.5}$ & $\textbf{15.0}$ & $\textbf{53.5}$ & $\textbf{23.1}$ \\
 \midrule
\multirow{2}{*}{\textbf{SParC}} & Qwen2.5-Coder & $67.3$ & $45.7$ & $69.0$ & $46.9$ \\
 & \quad + \ourmethod & $\textbf{68.4}$ & $\textbf{46.9}$ & $\textbf{69.6}$ & $\textbf{47.4}$ \\
 \midrule
\multirow{2}{*}{\textbf{CoSQL}} & Qwen2.5-Coder & $69.4$ & $40.3$ & $72.0$ & $41.3$ \\
 & \quad + \ourmethod & $\textbf{70.6}$ & $\textbf{42.3}$ & $\textbf{73.1}$ & $\textbf{42.7}$ \\
 \bottomrule
\end{tabular}
\caption{The main experimental results with and without \ourmethod. The best result under each setting is marked in \textbf{bold}.}
\label{tab:results}
\end{table*}


\subsection{Multi-Turn Text-to-SQL}
\subsubsection{Prompt}
This section aims to utilize the output from preprocessing to construct high-quality prompts (detailed in Appendix \ref{sec:appendix_prompt}), guiding the model in accurately generating SQL queries within multi-turn dialogue scenarios. Specifically, it includes: system prompts, which define the model's role, task, and output specifications; few-shot demonstrations, providing highly relevant references to help the model better understand query requirements; schema, which outline the database structure and relationships; and multi-turn dialogue, which leverage historical context to capture semantic associations and intent shifts, thereby improving query accuracy.

\subsubsection{Pre-SQL}
\label{sec:presql}
Considering that excessive table information in multi-turn dialogues may interfere with the model’s understanding of user intent, we first focus on filtering out table information that is irrelevant to the user’s query. At this stage, we use a prompt as input to guide the large language model in pre-generating an SQL query \cite{li_pet-sql_2024}. Subsequently, we refine the generated SQL query by eliminating unnecessary table information, ensuring that only relevant tables and fields are extracted. This process not only guarantees a high degree of alignment between the SQL query and user intent but also effectively reduces redundancy, thereby enhancing query accuracy and execution efficiency.

\subsubsection{Self-Debug}
\label{sec:debug}
Self-debug \cite{wang_dac_2024} refers to the process of detecting errors in the generated SQL query and then reintroducing the error information, along with table schema details and the user query, back into the model to facilitate error correction. This approach is inspired by the methodology presented in \cite{chen_teaching_2023}. During this process, the model leverages syntax error prompts, database schema information, and the original user query to generate a revised SQL query. By iteratively debugging itself, the model not only identifies and rectifies syntax errors but also improves its understanding of the query, thereby optimizing the SQL generation process.

\subsection{Presentation}
To enhance user experience, \ourmethod provides a transparent interaction mechanism, allowing users to clearly understand the SQL generation process and obtain real-time query results.
\paragraph{Inference Process Visualization}
The system provides a step-by-step explanation of the SQL generation and refinement process to help users better understand the query.
\paragraph{Real-time execution results}
SQL query results are displayed in tabular format, allowing users to quickly verify the accuracy of the generated SQL and enhancing the interactive experience.

\section{System Design}
This section presents the web design of Abacus-SQL to help users better understand the system’s features and how to interact with it.
\subsection{Frontend}
The front-end of \ourmethod (Figure \ref{fig:demo}) is built using Streamlit \cite{streamlit_streamlit_2024}, designed to provide a simple and intuitive user interface that enhances the overall user experience. As a comprehensive text-to-SQL system, \ourmethod incorporates a range of core functionalities, including:
\paragraph{User Authentication} Integrates a lightweight login system supporting account registration and encrypted password storage, along with Huozi \cite{huozi-team_hit-scirhuozi_2024} account login compatibility, ensuring privacy protection and seamless access.
\paragraph{Conversation Management} Supports multi-session management, allowing users to store query history and dialogue context, thereby enhancing interaction continuity and traceability.
\paragraph{Database Content Visualization} Provides an intuitive interface that clearly displays database tables, fields, and data, allowing users to easily browse and verify SQL queries.
\paragraph{Streaming output} Supports real-time streaming of the SQL generation process, reducing wait time and allowing users to access partial results earlier, thereby enhancing the interactive experience.

\subsection{Backend}
The backend of \ourmethod is built on FastAPI, providing efficient and flexible service capabilities while optimizing streaming output support. The backend utilizes Qwen2.5-Coder-7B \cite{hui_qwen25-coder_2024} for SQL generation. Although it has not undergone fine-tuning, its strong generative capabilities are sufficient for general text-to-SQL tasks.
Additionally, \ourmethod supports remote LLM API services (such as GPT-4o \cite{openai_gpt-4o_2024} and DeepSeek-R1 \cite{deepseek-ai_deepseek-r1_2025}), allowing users to securely integrate these models via API keys to generate more precise SQL queries.

\section{Experiment}
\subsection{Experiment Setup}
\paragraph{Dataset}
The \ourmethod multi-turn text-to-SQL evaluation benchmark is based on three datasets: Chase-C \cite{guo_chase_2021}, SParC \cite{yu_sparc_2019}, and CoSQL \cite{yu_cosql_2019}.
Chase is currently the largest cross-domain, context-dependent Chinese Text-to-SQL dataset. It consists of $5,459$ conversational turns ($17,940$ questions) spanning over 280 databases. Unlike other datasets, Chase-C features manually crafted questions based on database schemas from scratch, making it more realistic for practical applications.
SParC is a cross-domain, multi-turn Text-to-SQL English dataset. It comprises approximately $12,000+$ annotated natural language question-to-SQL pairs. These questions are derived from $200$ complex databases covering $138$ distinct domains.
CoSQL is another cross-domain, multi-turn Text-to-SQL English dataset. It contains over 3,000 conversational turns with 10,000+ annotated SQL queries. Each dialogue in CoSQL is specifically designed to simulate real-world database interaction scenarios.

\paragraph*{Metric}
To evaluate the performance of \ourmethod, we use two metrics: Question Execution Accuracy (QEX) and Interaction Execution Accuracy (IEX) \cite{zhang_coe-sql_2024}. QEX measures the execution accuracy of single-turn SQL queries, similar to EX, but focuses on the query result for individual questions. IEX assesses the execution correctness of all SQL queries across multiple interaction turns, ensuring that the system consistently generates accurate SQL throughout the entire conversation. Together, these metrics provide a comprehensive evaluation of the system's text-to-SQL capability in multi-turn dialogue scenarios.

\paragraph*{Model}
We used Qwen2.5-Coder 7B and 32B to evaluate the performance of \ourmethod on multi-turn text-to-SQL tasks. Qwen2.5-Coder \cite{hui_qwen25-coder_2024} is a code generation model based on Qwen2.5, equipped with powerful code understanding and generation capabilities. It is suitable for tasks across various programming languages, including SQL query generation. We set the inference to $3$-shot with a temperature of $0$.

\subsection{Main Result}
As shown in Table \ref{tab:results}, \ourmethod demonstrates improvements across all datasets compared to the baseline, with significant enhancement observed in the Chase-C dataset, highlighting its strong competitive edge in this domain. We also conducted ablation experiments on the pre-SQL and self-debug methods, finding that both approaches can improve system performance, with particularly more significant effects on Chinese datasets, thereby validating the effectiveness of the methods. (Appendix \ref{sec:appendix_ablation}). This result underscores \ourmethod's exceptional ability in multi-turn dialogue understanding and SQL generation, indicating its immense potential in applications that combine database querying and natural language processing.

\section{Conclusion}
We propose \ourmethod, a novel multi-turn dialogue-oriented text-to-SQL system designed to enhance database retrieval, cross-domain transferability, and SQL generation accuracy and efficiency. \ourmethod tackles existing challenges in current systems, such as the inability to efficiently retrieve relevant databases from open-domain database environment and the difficulty in transferring across diverse domains. By integrating the Murre method for efficient database retrieval, the Fused method to improve data generalization, and a combination of Pre-SQL and Self-debug to optimize query parsing, \ourmethod demonstrates exceptional adaptability and stability in handling complex query tasks. These results validate its effectiveness in real-world applications.


\bibliography{Abacus-SQL}

\appendix

\begin{table*}[t]
\small
\centering
\begin{tabular}{l|cccc|cccc|cccc}
\toprule
    \multirow{3}*{\textbf{Qwen2.5-Coder}} & \multicolumn{4}{c|}{\textbf{Chase-C}} & \multicolumn{4}{c|}{\textbf{SParC}} & \multicolumn{4}{c}{\textbf{CoSQL}} \\
    & \multicolumn{2}{c}{\textbf{QEX}} & \multicolumn{2}{c|}{\textbf{IEX}} & \multicolumn{2}{c}{\textbf{QEX}} & \multicolumn{2}{c|}{\textbf{IEX}} & \multicolumn{2}{c}{\textbf{QEX}} & \multicolumn{2}{c}{\textbf{IEX}} \\
    & \textbf{7B} & \textbf{32B} & \textbf{7B} & \textbf{32B} & \textbf{7B} & \textbf{32B} & \textbf{7B} & \textbf{32B} & \textbf{7B} & \textbf{32B} & \textbf{7B} & \textbf{32B} \\
\midrule
    \textbf{\ourmethod} & $45.5$ & $53.5$ & $15.0$ & $23.1$ & $68.4$ & $69.6$ & $46.9$ & $47.4$ & $70.6$ & $73.1$ & $42.3$ & $42.7$ \\
\midrule
    \textbf{\quad- Pre-SQL} & $45.5$ & $51.9$ & $14.3$ & $21.7$ & $67.1$ & $69.3$ & $45.5$ & $46.9$ & $70.4$ & $72.7$ & $41.3$ & $42.0$ \\
    $\quad\Delta$ & $-0.0$ & $-1.6$ & $-0.7$ & $-1.4$ & $-1.3$ & $-0.3$ & $-1.4$ & $-0.5$ & $-0.2$ & $-0.4$ & $-1.0$ & $-0.7$ \\
\midrule
    \textbf{\quad- Self-Debug} & $41.9$ & $48.7$ & $12.3$ & $19.8$ & $67.7$ & $69.1$ & $45.8$ & $47.1$ & $69.8$ & $72.2$ & $40.3$ & $42.0$ \\
    $\quad\Delta$ & $-3.6$ & $-4.8$ & $-2.7$ & $-3.3$ & $-0.7$ & $-0.5$ & $-1.1$ & $-0.3$ & $-0.8$ & $-0.9$ & $-2.0$ & $-0.7$ \\
\bottomrule
\end{tabular}
\caption{Ablation studies removing Pre-SQL or Self-Debug, The $\Delta$ row represents the differences with respect to \ourmethod.}
\label{tab:ablation}
\end{table*}

\begin{figure*}[t]
  \centering
    \includegraphics[width=\textwidth]{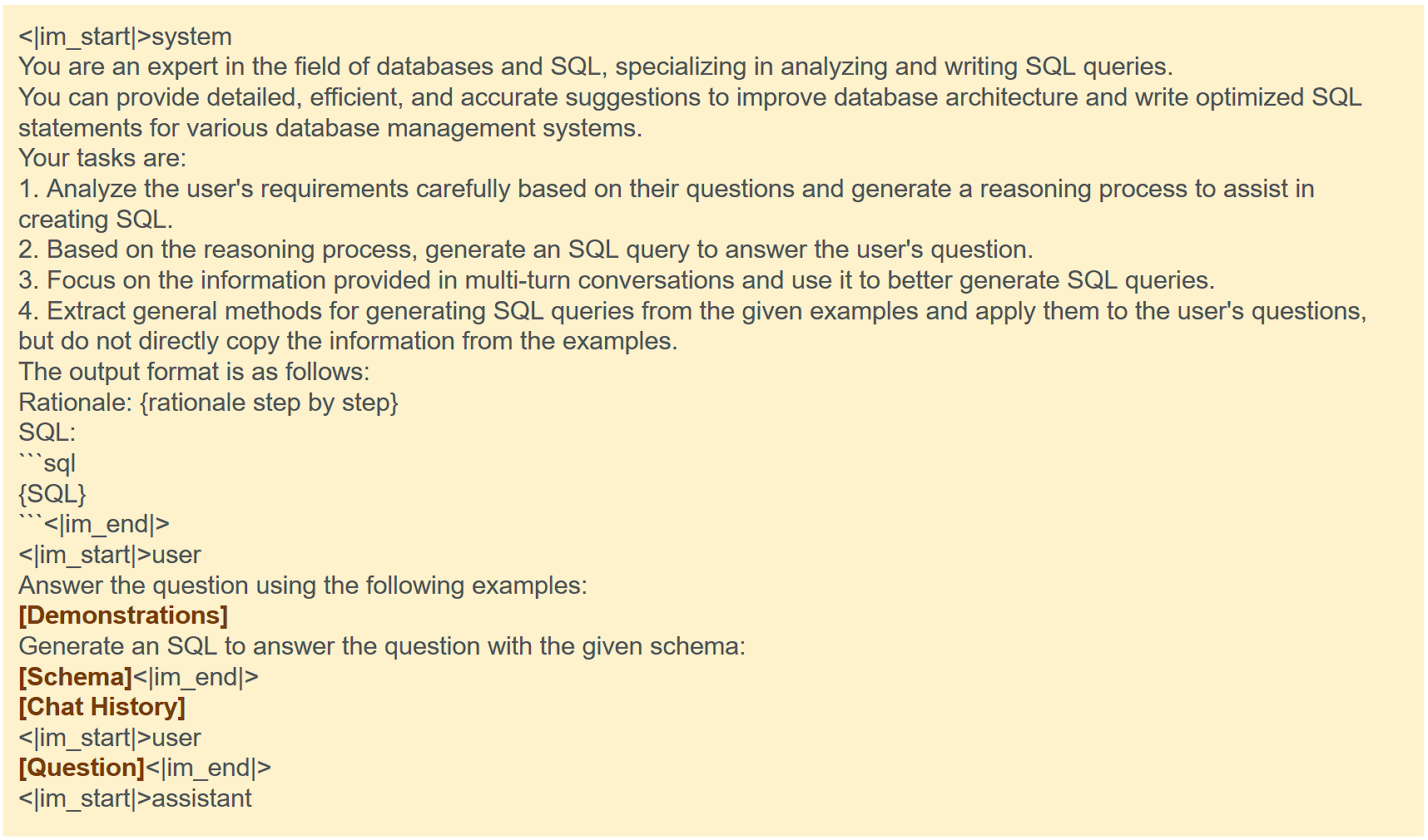}
    \caption{The prompt used in Multi-Turn text-to-SQL}
    \label{fig:prompt}
\end{figure*}

\section{Murre}
\label{sec:appendix_murre}
In terms of implementation, we adopted the multi-round retrieval method proposed by \cite{zhang_multi-hop_2024}, with the following steps:
\paragraph{Retrieval}
First, we extract table information from all databases in a multi-database environment and analyze the relevance of each table to the user's query. These tables are then ranked, and the top-k tables with the highest scores are retrieved.
\paragraph{Removal}
Next, a large language model (LLM) is used to rephrase the user's query, removing information related to the tables retrieved in the previous step. This step aims to eliminate tables that, although similar to the previously retrieved ones, are not relevant to the user's query.
\paragraph{Continue}
The retrieval process is then repeated from step $1$, continuing until all relevant tables from the related databases have been retrieved, ensuring comprehensive coverage of all information pertinent to the user's query.

By employing this multi-round retrieval and information removal strategy, \ourmethod can efficiently locate and extract the most relevant table information from the databases, thereby generating more accurate SQL queries.
\section{Fused}
\label{sec:appendix_fused}
Regarding the specific implementation of data augmentation, we adopted the FUSED method \cite{wang_improving_2024} to augment the dataset. Our specific implementation process is as follows:
\paragraph{User data upload}
The dataset uploaded by the user must include database schema and example SQL queries along with their corresponding natural language question descriptions. The system will validate the format of the uploaded data to ensure it meets the basic requirements for augmentation. If no user data is uploaded, the system will use a default dataset for demonstration augmentation.
\paragraph{Sample sampling and clustering}
The system clusters the demonstrations based on structural features of the SQL queries (such as keywords, operators, etc.), forming different semantic categories. It then randomly samples demonstrations from each category, ensuring that the demonstrations input into the augmentation process exhibit significant diversity, thus avoiding overly similar demonstrations.
\paragraph{Sample fusion}
Using a large language model (LLM), the sampled demonstrations are used as inputs to generate new demonstrations through few-shot learning. The newly generated demonstrations combine features from multiple demonstrations while maintaining differences from existing ones, thereby enhancing the diversity of the overall demonstration pool.
\paragraph{Verification and filtering}
The system performs semantic consistency verification on the generated SQL queries and question descriptions, ensuring that the generated demonstrations are consistent with the database schema and the intended query. Low-quality or redundant demonstrations are removed through automated testing.
\paragraph{Demonstration pool update}
The augmented dataset is automatically added to the demonstration pool and merged with the existing dataset. The merged Demonstration pool is used for subsequent model inference and training, further improving the accuracy and adaptability of the generated SQL queries.

\section{Prompt}
\label{sec:appendix_prompt}
The prompt for \ourmethod, shown in Figure \ref{fig:prompt}, mainly consists of the following components: system prompts, few-shot examples, schema, and multi-turn dialogue.

\section{Ablation Studies}
\label{sec:appendix_ablation}
As shown in Table \ref{tab:ablation}, we conduct ablation experiments on Pre-SQL and Self-debug methods, drawing the following conclusions:
\paragraph{Both methods improve system performance.} Pre-SQL reduces the interference of irrelevant tables, decreasing complexity and improving query efficiency. Self-debug addresses post-generation errors, reducing mistakes caused by input ambiguity or understanding bias, further optimizing accuracy.
\paragraph{The results are particularly significant on Chinese datasets.} Experiments show that when testing the Chase-C dataset on the Qwen2.5-Coder 32b model, Pre-SQL improves by $1.4$ points on the IEX metric, while the Self-Debug method enhances the IEX metric by $5.1$ points. Due to the ambiguity and complexity of the Chinese language, the system's semantic understanding requirements are higher. Pre-SQL helps reduce interference from irrelevant information, while the Self-Debug method corrects understanding biases. The synergy between these two methods significantly improves query accuracy and reliability, demonstrating a distinct advantage in handling Chinese natural language queries.

\end{document}